\documentclass[conference]{IEEEtran}

\usepackage[T1]{fontenc}
\usepackage[utf8]{inputenc}
\usepackage{amsmath,amssymb}
\usepackage{booktabs}      %
\usepackage{graphicx}
\graphicspath{{figures/}}
\usepackage{siunitx}       %
\usepackage{xcolor}
\usepackage[hidelinks]{hyperref}
\usepackage[capitalise]{cleveref}      %
\usepackage{comment}       %

\newcommand{\byterl}{ByteRL}
\newcommand{\locm}{LoCM}
\newcommand{\locmver}{LoCM~1.5}

\newcommand{\CertStrict}{51.35\%}        %
\newcommand{\CertStrictCI}{[50.37, 52.33]}
\newcommand{\CertClone}{60.4\%}          %
\newcommand{\NoSearchNew}{26.8\%}        %
\newcommand{\SearchGainNew}{24.6}        %
\newcommand{\AttackByteRL}{90.1\%}       %
\newcommand{\EngineSpeed}{443$\times$}   %
\newcommand{\DisambigStartWorlds}{$\sim$$2^{101}$}%
\newcommand{\PeakMem}{$\sim$0.9\,GB}       %
\newcommand{\CompMemCap}{256\,MB}          %
\newcommand{\ByteRLPeakMem}{$\sim$37\,MB}  %
\newcommand{\NoSearchMem}{$\sim$535\,MB}   %

\newif\ifshownotes
\shownotesfalse          %

\ifshownotes
  \newcommand{\todo}[1]{\textcolor{red}{[TODO: #1]}}
  \newcommand{\draftnote}[1]{\textcolor{blue}{[#1]}}
  \newcommand{\dustin}[1]{\textcolor{orange}{[Dustin: #1]}}
  \newcommand{\claude}[1]{\textcolor{teal}{[Claude: #1]}}
  
\else
  \newcommand{\todo}[1]{}
  \newcommand{\draftnote}[1]{}
  \newcommand{\dustin}[1]{}
  \newcommand{\claude}[1]{}
  \excludecomment{claudenotes}
\fi

\begin{document}

\title{Unsound Search with Policy and Value Networks\\in Legends of Code and Magic}

\author{%
  \IEEEauthorblockN{Dustin Rubin}
  \IEEEauthorblockA{\textit{Georgia Institute of Technology}\\
  Atlanta, Georgia, USA\\
  dustin.rubin5050@gmail.com} %
}

\maketitle

\begin{abstract}

Decision-time search in perfect and imperfect information games with enumerable belief states are effective methods for game AI.
Collectible card games are imperfect information games with large belief states.
Legends of Code and Magic is a collectible card game competition where the belief states are \DisambigStartWorlds{}.
The Legends of Code and Magic (\locm{}) champion, \byterl{}, plays with no search.
Other works claim sound enumeration-based search is unusable in the genre due to the number of belief states.
We measured three previously defined properties that predict where theoretically unsound perfect information Monte Carlo's defects are cheap and found \locm{} sits in the favorable region.
Starting with imitation learning of the runner-up policy, NeteaseOPD, we created a policy and value feed-forward network.
Our agent searches over worlds sampled from a prior over the opponent's deck built from the runner-up's drafts.
Using our strictest configuration in the battle phase we beat \byterl{} with a win percentage of \CertStrict{} 95{\%} CI \CertStrictCI{}, over 10{,}000 pre-registered games using the \locm{} official referee and time limit.
Search is not a minor factor on the matchup between our agent and \byterl{}.
Without search this agent scores \NoSearchNew{} and adding search adds $+$\SearchGainNew{} points.
Unsound search in imperfect information games could be exploitable.
We replicate a published best-response attack against \byterl{}.
We then apply the same attack protocol to two search configurations of our agent, and each one resists it better than \byterl{} at every iteration.
In \locm{} unsound search gives us a stronger and more resilient agent.

\end{abstract}

\section{Introduction}

\begin{figure}[!t]
  \centering
  \includegraphics[width=\columnwidth]{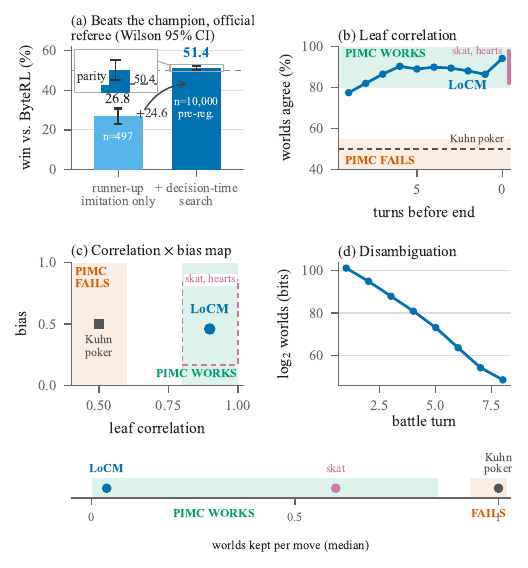}
  \caption{Figure A shows the uplift of decision time search on our imitation learned NeteaseOPD policy certified on the official referee within the time limit. a is the certified win rate of our most restrictive agent vs ByteRL with and without search with the zoom in showing the CI is above 50\%. The other diagrams show the empirical factors that support PIMC working well in LoCM.
  Disambiguation shows the number of consistent states shrinking as actions are taken, leaf correlation shows determinizations not flipping the outcome and also shows minimal bias.}
  \label{fig:headline}
\end{figure}
In collectible card games (CCGs) players construct a deck of game
pieces and draw from it as a shuffled pile. They are imperfect information games where players keep their hand
and deck contents secret. CCGs also have massive state spaces as hand contents and deck order create
a combinatorial number of states. Player actions can leak hidden information about their hand.

Search at both training and decision time has been used in work that resulted in superhuman performance in perfect
information games like Go~\cite{silver2016alphago} and imperfect information games like poker~\cite{moravcik2017deepstack}.

Legends of Code and Magic (\locm{}) is a CCG designed to include many of the challenges of
CCGs but also simplify building AI agents. In its 1.5 version it involves a pool of
120 randomly generated cards. Then players must construct a 30 card deck with no more than 2 copies of
any card~\cite{kowalski2023locm}.
This emulates players creating and battling in the constructed format but with a card pool that cannot
be memorized. In this paper we evaluate the battle phase of this competition in
isolation. Deck construction is fixed and identical for both
players (\cref{sec:setup}). Even though we are fixing the deck, one axis we study is how much information the agent assumes about the opponent's
deck list. Even when the agent knows the opponent's list, as in many competitive CCG tournaments, hand and
draw order still create a combinatorial state space.
The leading policy in \locm{} is \byterl{}~\cite{xi2023byterl}, which chooses to abandon decision-time search using only a fixed policy with a recurrent neural network. It also does not use training time search as they found that it did not
improve the policy versus spending more time generating games. This is due to the limitations of sample throughput of the game simulation during training. At decision time \byterl{} is purely a single evaluation of the history of actions and the current state.

Current leading sound search algorithms for imperfect information
games are ReBeL \cite{brown2020rebel} and Student of Games
\cite{schmid2023sog}. Both require
enumeration of states, which proves intractable in CCGs. Sound sampling methods such as Monte Carlo CFR~\cite{lanctot2009mccfr,lisy2015oos} are too
slow to converge with high numbers of states.

Perfect information Monte Carlo (PIMC)~\cite{long2010pimc} is also a common approach to use for agents working on imperfect information games.
It is known to be theoretically unsound due to strategy fusion and non-locality~\cite{frank1998search}, yet it has powered effective play in
trick-taking games~\cite{ginsberg2001gib,buro2009skat}. Long et~al.~\cite{long2010pimc} showed empirically that in
certain game structures these algorithm defects often do not affect correct play.

We measured \locm{} game tree using those empirical properties (\cref{fig:headline}b--d) and found that leaf correlation is high, bias is near .5, and disambiguation is strong. These three factors mean that theoretically unsound search on this CCG is similar to other games where these defects do not affect correct play. This gives us some empirical support and justification for exploring this type of search on this game and potentially more broadly across CCGs where leading sound search algorithms are not effective.
We build an engine that is orders of magnitude faster at evaluating moves than gym-locm. Also rather than training new models we use imitation trained policies utilizing feed forward policy and value networks of both the runner up and top models from the competition using supervised learning on their self play games to be able to have a starting model similar in performance to those competitors. Our engine also supports efficient sampling of consistent worlds utilizing a prior filtered by the game rules. This method is supported by a partial filter ensuring sampling from consistent worlds instead of using a recurrent neural network over time.  Utilizing these tools we measure the performance enhancements utilizing sampled search over possible worlds and show how a network imitating the runner up with theoretically unsound search beats the champion within the official time limit (\cref{fig:headline}a). We also show how much stronger the leading policy can be utilizing this search.
On perfect information games exploitability has been shown to be harder against agents utilizing search~\cite{wang2023adversarial}. One potential downside of utilizing a theoretically unsound search is that the final policy could be stronger in a vacuum but more exploitable. To test the exploitability of this new agent we replicate a published best response attack against \byterl{} and then use that same best response attack against our new policy and show that our policy is more resistant to this attack at all iterations.

\section{Related Work}

\subsection{Agents for Legends of Code and Magic}
LoCM was built for AI competitions \cite{kowalski2023locm}, cards are procedurally generated per game so agents must generalize over card properties. Classic search won competitions under the fixed 1.2 rules. Learning agents took over in 1.5, the version we are exploring.
gym-locm gave RL baselines for the battle phase \cite{vieira2023gymlocm}. \byterl{} \cite{xi2023byterl} won both the COG 2022 tracks using industrial scale training and no search at inference. Halu\v{s}ka and Schmid \cite{haluska2024exploitability} beat it with a win rate of $\sim$90\% with a best-response attack.

\subsection{Sound Search in Imperfect-Information Games}
Extending sound search to imperfect information games has been pursued by DeepStack \cite{moravcik2017deepstack} using CFR \cite{zinkevich2008cfr}, and with alpha zero like self-play training by ReBeL \cite{brown2020rebel}, and Student of games \cite{schmid2023sog}.
These all resolve subgames over explicit belief distributions. All of these require their belief states to be enumerable, such as hold 'em poker's 1,326 possible pairs. A CCG's hidden state is combinatorially large.
In LoCM \DisambigStartWorlds{} at battle start is too large to enumerate. Prior LoCM work makes the infeasibility claim explicit saying enumeration based search ``rendering the current search methods unusable''
\cite{haluska2024exploitability}. \byterl{}'s reasoning for lack of training-time search is different and is based on sample throughput \cite{xi2023byterl}.

\subsection{Determinized Search}
PIMC \cite{long2010pimc} first determinizes into W worlds and then searches each with perfect-information methods and then aggregates. It has two main defects, strategy fusion and non-locality \cite{frank1998search}.
However, it still is able to give strong play in bridge \cite{ginsberg2001gib} and skat \cite{buro2009skat}. Long et al. measured which game structures make the defects not matter to performance. We measure
those structures for LoCM. Determinized search reached CCGs before in MtG \cite{cowling2012mtg}, ISMCTS \cite{cowling2012ismcts}, Hearthstone \cite{santos2017hearthstone,swiechowski2018hearthstone}, LoCM bots \cite{cao2025locm}.
None of these were evaluated against a scaled-RL state of the art agent. For robustness in Go, search partly resists adversarial policies \cite{wang2023adversarial}. Poker's LBR probe \cite{lisy2017lbr} vs DeepStack
is the closest imperfect-info evidence.

\subsection{Relation to Prior Work}
Versus sound search, sampled belief does not enumerate all consistent states and thus operates where sound search cannot. Versus the exploitability study we replicate and explore how our strategy increases search defenses.

\section{Method}\label{sec:method}

First we reimplement \locm{} rules and RL training env in a new Rust engine.
Using LLM based coding and extensive testing creation of this engine is automated. 
The new engine has a few properties, first is speed. The engine simulates 
\EngineSpeed{} more games per second than the reference gym-locm implementation. This makes 
search more feasible at runtime within the tight per move runtime but also allows 
use of training imitation policies of the leading policies on cheap hardware. The 
second utility of this new engine is its ability to do efficient determinization
with full control over the distribution of randomness. This allows the agents to 
eliminate known impossible states as well as use a prior to predict the opponent's hand.

The headline claims against \byterl{} do not depend on the correctness of this engine. All certified numbers 
were run on the official competition engine with referee, our agent received an isolated process
where it uses this reimplemented engine to search using only the information it has available.

This engine is still used for several of the other tests and to ensure correctness extensive
testing is done comparing that games play exactly the same.

\subsection{Observation and Action Encoding}
We adopt the \byterl{} encoding of the state and actions. Sharing this simplifies behavior cloning of the algorithm by
letting us do supervised learning over self-play games using our engine. We use the same per-card atoms and action space as \byterl{}. Our nets are feed-forward over the current state rather than an LSTM like \byterl{}. Our 
networks only ever see their own player's cards. 
\subsection{Search}
Since \locm{} players complete an entire turn before passing back to their opponent we search possible turns using beam search. The search ends when the plans end in a pass. Next we sample W worlds from a prior over the opponent's remaining 
hand and deck. The opponent's exact deck is not known so we test various possible distributions. Assuming uniform over the available draftable pool, picks based on rivals picks from the pool (alone or mixed 50/50 with uniform), picks based on assuming the opponent chose the same cards as us,
or the exact decklist provided by the opponent (this is often common in commercial CCG competitions). Next we roll out one opponent reply turn and score the resulting state with the value head. We average scores across worlds and the plan with the 
highest average is executed. When the clock is in use worlds are evaluated in chunks while still under a 185 ms budget, a safety margin under the 200 ms timer. This is essentially PIMC with the policy and state valuation coming from learned networks.
The certified agents runs beam 24, seven expansions, and worlds up to 384 under the clock. 

\subsection{Training}
For training all networks trained are dual-head with a policy and value and are initialized by behavior cloning. Using our engine we have the teacher create a self-play corpora, then we train both heads jointly. We use cross-entropy on the action 
chosen by the teacher to train the policy head. We use regression on the value head to predict the game outcome. We train two main lines, one based off of \byterl{} self-play and a line that includes no play from \byterl{} taught by the runner 
up NeteaseOPD and it is kept strictly separate. Data and network capacity are varied on a grid. The resulting networks without search recover most of the teacher strength unaided simply greedily using the policy head. All of our search agents in this paper are one of these networks plus the 
decision time search. None of our headline agents are trained or fine-tuned against \byterl{}. The best-response attackers are trained against frozen defenses.

\section{Experimental Setup}\label{sec:setup}

\subsection{Game and Protocol}
The rules are \locmver{} under the COG 2022 competition rules. We are investigating 
just the battle phase, so for each round we have \byterl{} draft cards for both 
players. Both players enter battle with identical decks and then the battle 
phase decides the game. The player that goes first alternates between games, the 
seeds that are chosen are disjoint across all tests.

There are two modes of play we measure. The first mode is the internal harness with no clock to do ablations of
play quality and different search depth. The other we use the official referee, 200 ms move timer after the first move. In this mode 
our agent runs in an isolated process and talks to the official referee using the competition protocol. Our 
agent is given only partially observable data and uses that to construct the game using our engine and 
must give a response within the move timer.

We call a result "certified" when we are using the mode with the official referee and timer with 
predefined seeds.
For all certified runs we use an AWS g5.2xlarge (A10G, 8vCPU, 32 GB). \locm{} did
not publish a hardware spec but this is an isolated mid spec cloud computer. 
Hardware speed can affect how much runtime search is possible and thus slightly different
hardware configurations could improve or hurt our agent. For non certified runs 
hardware is more flexible as execution speed was not part of the evaluation. How 
much budget was available for certified runs depended on hardware strength.

\subsection{Corpora}
To generate play data that we use for imitation learning of the top policies first 
we use our engine mode along with their play to generate 80k games of \byterl{} self play (36k of which are used for training)
and 126k games of NeteaseOPD self play. NeteaseOPD is the COG 2022 runner-up. This is 
run as a self play protocol using its published policy. We also use this policy as a way 
to generate a belief prior \cref{sec:method} for actual play. For early self-play corpora
we use temperature-sampled decisions, these are excluded from value targets. 

\subsection{Certification}
To ensure certification of games was fair we use a cloud compute instance to make sure 
both agents can perform without interruption. To ensure we are never violating 
the 200 ms timer we ensure that all moves are run within the per move time limit 
and that no move exceeds the timer. All certified results are in disjoint seed blocks 
to ensure no unfairness from seeds. Win rates at 1000 games could move $\pm3$ points under 
these tests which is a large amount of variance. Using the approximated win rate without 
the official judge we determined how many games would be needed to have a confident result.
Every interval is a Wilson 95\% CI. Before running certification the config, number of games 
and seed block were chosen. 

\subsection{Attack Protocol}
Following the protocol of \cite{haluska2024exploitability}, the attacker is a policy fine-tuned against 
the frozen defense on a fixed pool of 32 decks. The search defenses are run with the opponent's list known. To make the attacks as strong as possible every attack runs 
to convergence with a plateau verified by oscillation. We report the win rate as mean of the last 5 along with a peak.
We replicated this against \byterl{} and received a peak win rate of 97.0\% and a similar replicated converged rate of \AttackByteRL{} to the paper's 90.4\%.

\subsection{Game-Structure Measurement}
To test factors with Long et al. parameter measurement in \cref{fig:headline} we do self play with the NeteaseOPD trained net with no search. We use 32 fixed decks with 600 games.
For disambiguation we measure the log2 count of consistent opponent hidden states. For leaf correlation, we sample two consistent worlds under uniform belief and play each out and if there is the same winner they have more leaf correlation. 
For bias we check how often the first player is the winner of those near terminal outcomes across the sampled worlds, with 0.5 meaning no lean.

\section{Results}\label{sec:results}

\subsection{Certified Results Against \byterl{}}
With knowledge of the opponent's deck and the policy trained from \byterl{} known search defeats the champion with a win rate of \CertClone{} using 500 certified games. To perform our strictest test we use our NeteaseOPD self play and remove all 
knowledge of the opponent's deck using the NeteaseOPD prior mixed 50/50 with a uniform distribution and it still certifies at \CertStrict{} \CertStrictCI{} over 10{,}000 games. Note that this number of games was precomputed based on estimated win 
rate of the players and was a fixed number of games that were intended to be played. The margin is created by the search. The same network playing policy argmax certifies at \NoSearchNew{} so search is worth \SearchGainNew{} points. Note there is 
first player advantage in the matchup, the agent wins 54--56\% going first and 46.6\% going second but results were made with an even amount of both players going first. 

\subsection{Search Budget Ablations}
\begin{figure}[t]
  \centering
  \includegraphics[width=\columnwidth]{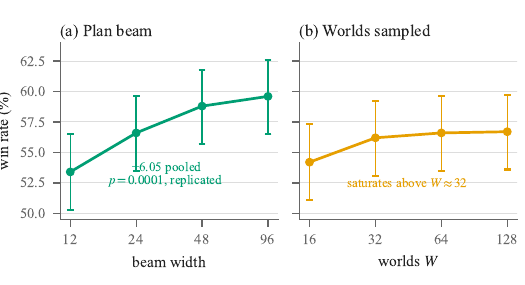}
  \caption{Here we explore varying (a) plan beam width and (b) worlds sampled as beam search raises it continues to improve. Worlds sampled saturates at around 32.}
  \label{fig:budget-ladder}
\end{figure}

We test various configurations of search to see what is most effective. We find that candidate-plan beam size from 12 to 96 continues to increase strength. We find that beyond 32 worlds adding more does not increase performance. More 
depth does not increase agent quality. Most of the value of the search is immediately gained by evaluating plans. This leads to the idea that coming up with a full plan for the turn and evaluating the result 
is one of the main driving factors of the increased performance. 

\subsection{Belief About the Opponent Deck}
If our agent knows the opponent's exact list it gives a clear advantage to our search. With a uniform assumption we achieve a win rate of 50.5. We get 56.6 with our agents having the runner-up prior.
If we know their exact list we get 59.6.
\subsection{Training Data and Model Size}
We also measure how much raising model size and data does on performance of our agents. What we find is that quadrupling training data moves win rate 8 points for both a 2.5M and 10M parameter model. Model size does 
not seem to be a contributing factor to performance at this level of games and for imitation learning. This means that a smaller model that can run faster and more parallelizable may allow a larger search budget under
time constraints. Validation loss does show capacity x data interaction but none of it converts to wins.

\subsection{Exploitability}

We first replicate an attack against \byterl{} and find that adding search cuts converged exploitability by 21 to 28 points. Note that for exploitability testing we do not use the official certification budget 
and instead use $W{=}16/64$ this is less search than can be achieved in the time limit in the certification. The certification resistance to exploitability is unmeasured. This shows that against this attack this unsound search reduces exploitability.

\begin{table}[t]
  \centering
  \caption{Converged exploitability (250-iteration PPO best
  response, this is the mean of last five evaluations) and
  head-to-head strength. The head-to-head numbers are internal runs
  at $W{=}64$ for both rows, with the uniform belief for the first
  and the runner-up prior for the second. The exploitability runs
  use $W{=}16$ and $W{=}64$ with the opponent's list known, so
  compare within a column and not across.}
  \label{tab:exploit}
  \begin{tabular}{lccc}
    \toprule
    Defense & h2h vs.\ \byterl{} & converged & peak \\
    \midrule
    \byterl{} (no search) & $\sim$50 & 90.1 & 97.0 \\
    plan search, $W{=}16$ & 50.5 & \textbf{61.9} & 65.0 \\
    plan search, $W{=}64$ & 56.6 & 69.3 & 74.5 \\
    \bottomrule
  \end{tabular}

\end{table}

\begin{figure}[t]
  \centering
  \includegraphics[width=\columnwidth]{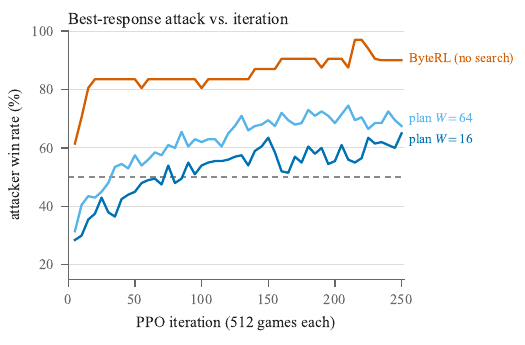}
  \caption{Best-response attack win rate over 250 PPO iterations and the three defenders. The attack on \byterl{} reaches 90\% within 50 iterations, while both search configurations hold the attacker. The defenses run at $W{=}16$/$64$ which is far below the certified budget.}
  \label{fig:exploit-strength}
\end{figure}

\section{Discussion and Future Work}\label{sec:discussion}

\subsection{Game Structure and Determinized Search}
Long et al.'s three properties predict PIMC works, we measure them and find that all three support this game being one where PIMC works. Disambiguation is strong, as players play the cards in their hand
the number of consistent belief states shrinks dramatically. This is likely consistent across CCGs. Bias is near neutral. Leaf correlation is high supporting that consistent worlds tend to agree on a winner, this may 
be because most games are not decided by a showdown of having or not having a particular card in your hand, but this could be different across CCGs. World samples also saturate at just 32 implying that a sample of that size 
is already predictive of good strategy and with leaf correlation where it is is a good guide for action selection.

\subsection{Decision-Time Compute Under the Competition Limits}
The competition capped decision-time resources (200 ms clock, 256 MB RAM) but not training compute. The incumbent's main advantage is scaling on an uncapped axis. 
We show just how valuable decision time compute can be in this type of competition. On memory our agent peaks at \PeakMem{} and \NoSearchMem{} with no search at all. We exceed the \CompMemCap{} and the champion 
does not. This is a real limitation but the memory cost of our inference runtime (libtorch) alone exceeds the cap before any search runs, and the champion peaks at \ByteRLPeakMem{}. It is possible that optimizations could bring this memory usage down. 
It is worth noting that this competition was played on unspecified hardware and the hardware choice may not have considered the combination of parallel runtime search utilizing GPU inference. The clock 
which we do consider we compete fairly within the competition's limit. In terms of having the strongest agent runtime search on top of the existing training runs provides a clear advantage.

\subsection{Imitation Fidelity, Head-to-Head Strength, and Robustness}
Imitation fidelity does not directly translate to playing strength. This could also be due to the value net
saturating and that combination of a saturated value net plus plan search may be the primary driver of performance even if the policy has more to learn.
Head-to-head playing against other agents does not translate to robustness against the best response attacks. We see that agents that are closely matched are 28 points apart against a best response learner. 
Our unsound search buys that even for an even policy. 

\subsection{Measuring a Small Margin}
Measuring the minor edge our agent has against \byterl{} required a large amount of games as the actual win rate has a $\pm$3 over 1000 games. To do this in a valid way 
we created the certification protocol with a pre-registered number of games and based on estimated win rate ran large competitions between the agents. 

\subsection{Limitations}
We focus solely on the battle phase without considering the draft phase. Our memory exceeds the competition cap and so this entry could not be an official competition entry. It is also unclear if the official 
competitions supported GPUs which we rely on for the parallelism needed to run our algorithm within the time limit. Based on the entrants it is possible that all neural network inference ran on CPU and thus 
we would hit limitations on hardware. Exploitability was one run per defense. 
The certified agent's exploitability is not measured and likely higher. Our structured PIMC efficacy measurements focused on just 32 decks.

\subsection{Future Work}
For future work moving this algorithm to see if it generalizes to stronger play in other CCGs. Also testing out sampled sound variants of search to evaluate if higher quality play could be achieved. 
\section{Conclusion}

Prior work concluded search is infeasible in CCGs because of belief enumeration. State of the art plays without it, sampling, while unsound empirically adds generic win rate strength to these agents.  Determinized search, PIMC, over consistency filtered beliefs beats the state of the art agent \byterl{} under the official referee and clock, \CertClone{}, with the opponent's deck known. \CertStrict{} shows our agent's final win rate 
over 10,000 pre-registered games without utilizing any of \byterl{} data. The margin comes from enabling search, +\SearchGainNew{} points over the network's raw policy. That same decision-time compute buys robustness. The 
attack that beats \byterl{} at \AttackByteRL{} but converges to 62--69\% against both search agents.

All of the evaluation choices tilt towards \byterl{}. We allow it to draft both decks so synergy with its policy and deck could give it an advantage. We never train against \byterl{} or condition on its play, our belief prior comes from the runner-up's drafting choice, and the clock caps exactly the resource our agent converts into playing strength. 

A rules engine gives our agent an advantage in that a policy must spend time approximating the consequences of card interactions and the set of hidden worlds still consistent with play and supplies both at decision time. 
LoCM's structure is measurably where PIMC defects are cheap. Unsound search at decision time over sampled beliefs is a practical way to improve agent strength and robustness in a domain where sound search cannot extend to.  

\section*{Acknowledgments}
Utilized AI assistance with coding, review and minor grammar and spelling of writing. 

\bibliographystyle{IEEEtran}
\bibliography{refs}

\end{document}